\documentclass{article} 
\usepackage{nips10submit_e,times}

\usepackage{graphicx} 
\usepackage{epstopdf} 
\usepackage{chngpage}
\usepackage{theorem}
\usepackage{subfigure}
\usepackage{amsmath}
\usepackage{amsfonts}
\usepackage{bm}
\newtheorem{theorem}{Theorem}

\newtheorem{proposition}{Proposition}

\newcommand{\E}{\mathbb{E}}
\newcommand{\prob}{\mathbb{P}}
\newcommand{\risk}{\mathcal{R}}
\newcommand{\Hyp}{\mathcal{H}}
\newcommand{\R}{\mathbb{R}}
\newcommand{\N}{\mathbb{N}}
\newcommand{\X}{\mathbb{Z}}
\newcommand{\Y}{\mathbb{Y}}

\DeclareMathOperator*{\argmin}{\arg\min}

\DeclareMathOperator*{\KL}{KL}

\DeclareMathOperator*{\entropy}{\mathbf{h}}

\title{Pair-Wise Cluster Analysis}

\author{
David R. Hardoon\thanks{www.davidroihardoon.com} \\
Department of Computer Science\\
University College London\\
London\\Ä
\texttt{davidrh@me.com} \\
\And
Kristiaan Pelcksman\thanks{http://www.it.uu.se/katalog/kripe367} \\
Division of Systems and Control\\
Dept. of Information Technology
Uppsala University, Sweden \\
\texttt{kristiaan.pelckmans@it.uu.se}
}

\nipsfinalcopy 

\begin{document}

\maketitle

\begin{abstract}
This paper studies the problem of learning clusters which are consistently present in different (continuously valued) representations of observed data.  Our setup differs slightly from the standard approach of (co-) clustering as we use the fact that some form of `labeling' becomes available in this setup:  a cluster is only interesting if it has a counterpart in the alternative representation. The contribution of this paper is twofold:  (i) the problem setting is explored and an analysis in terms of the PAC-Bayesian theorem is presented, (ii) a practical kernel-based algorithm is derived exploiting the inherent relation to Canonical Correlation Analysis (CCA), as well as its extension to multiple views. A content based information retrieval (CBIR) case study is presented on the multi-lingual aligned Europal document dataset which supports the above findings.
\end{abstract}

\section{Introduction}

\begin{figure}
\label{learning}
\begin{minipage}[b]{0.5\linewidth} 
\centering\includegraphics[width=5.5cm]{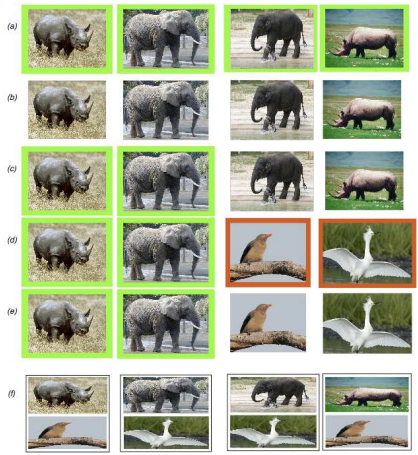}
\end{minipage}
\hspace{0.5cm} 
\begin{minipage}[b]{0.5\linewidth}{\caption{
		Pictorial representation of different learning paradigms, extending the picture in \cite{raina2007self}.
		Suppose the aim is to discriminate elephants from rhinos.
		When a picture appears in a frame, a corresponding class-label is available.
		In cases:
		(a) supervised classification.
		(b) unsupervised learning.
		(c) semi-supervised learning.
		(d) transfer learning (the two different colors indicate two different learning tasks).
		(e) selftaught learning, and 
		(f) pairwise cluster analysis (PWCA).
		Note that in the latter we try not to find the class labels themselves, but to recover the 
		symbiotic relation between elephant-egret, and rhino-oxpeckers.
		Specifically, the presence of oxpeckers might help us in predicting the presence of a rhino, and vice versa.}}
\end{minipage}
\end{figure}

Consider the setup where individual observations come in two different representations $(x,y)$. This paper focuses on the questions: `If we observe a new $x$, what can be said about the corresponding $y$, and vice versa?' While this abstract problem has obvious relations to classical supervised learning, its inherent symmetry relates it to unsupervised learning as well. This paper studies the above problem, specifying the properties to be predicted in terms of pre-specified membership functions. Figure (\ref{learning}) differentiates the above problem - termed PairWise Cluster Analysis (PWCA) - from the supervised, unsupervised, semi-supervised, transfer- and multiple-task learning \cite{bendavid2003} and self-taught learning \cite{raina2007self}. The present learning strategy has direct relations to co-occurrence analysis, co-clustering \cite{banerjee2005clustering}, kernel Canonical Correlation Analysis (kCCA) \cite{Hardoon04} and has been motivated by the previous works of
  Pelckmans et al. \cite{pelckmans06mc} and Sim et al. \cite{kelvin} which explore an application in relating text corpus - microarray expression and multi-attribute co-clustering respectively.

The analysis given in Section 2 phrases the learning problem in terms of the PAC-Bayesian theorem, much in the spirit of the recent work of Seldin \& Tishby \cite{seldin2009pac}. Although, while the latter concerns density estimation for discrete variables,  the presented ideas cover a spectrum of unsupervised learning (clustering).  The analysis presented in \cite{seldin2009pac} concerns, essentially, the same quantity $E_Q[\risk(h)]$ as in subsection 2.1, equation (\ref{eq.explanatory}), which characterizes how well some hypotheses $Q$ aligns with the distribution underlying the data. Our extension to pairwise clustering is fundamentally different - incorporating a notion of prediction  `loss' - while the relation of Kuller-Leibler ($\KL$) divergence and the norm of an hypotheses establishes a relation with the learning algorithm.

Section 3 (i) derives an effective learning algorithm, boiling down to a quadratic (or a generalized) eigenvalue problem.
This learning machine is closely related to kernel Canonical Correlation Analysis (see e.g. \cite{BachJordan,Hardoon04} and references therein). Empirical (ii) evidence for this learning paradigm, and the proposed algorithm is then presented. We proceed to demonstrated the benefit of learning structure within the data on a multi-lingual text-corpora \cite{Koehn}. Section 4 indicates a number of open questions.

\section{A Generic Analysis using the  PAC-Bayes Theorem}

Consider a function $h_r:\{x\}\rightarrow[0,1]$ that verifies, for a given problem setting, 
how good a certain `rule' $r$ performs on a sample $x$. 
The goal of a learning algorithm is to find the best rule $r$ in a given set of plausible rules (the hypothesis set).
Then, learning proceeds by collecting a dataset $\{X_i\}_{i=1}^n$ of $n$ observations assumed to be sampled 
independently from identical distributions (i.i.d)\footnote{We will use the convention to denote stochastic variables 
as capital letters, e.g. $X,Y,\dots$,
while deterministic quantities are denoted in lower case, e.g. $h,f,i,x,y,n,\dots$.}.
The empirical risk $\risk_n(h_r)$ and the actual risk $\risk(h_r)$ of an `hypothesis' $h_r\in\Hyp$ is defined as
\begin{equation}
	\begin{cases}
		\risk_n(h_r) = \frac{1}{n} \sum_{i=1}^n h_r(X_i)\\
		\risk(h_r) = \E[ h_r(X)],		
	\end{cases}
	\label{eq.risk}
\end{equation}
where the expectation $\E[\cdot]$ concerns the fixed, unknown distribution underlying the $n$ i.i.d observations. 
For supervised learning problems, (informally) an observation $x$ consists typically of a couple $(z,y)$ with 
a covariate $z$ and an `output' $y$.
Then $h_r$ is often rephrased as $h_r(x)=\ell(y-r(z))$, 
where $\ell:\R\rightarrow[0,1]$ is the `prediction loss' between the actual observation $y$ and its prediction $r(z)$.
In a Bayesian context, we assume that the hypothesis $h_r\in\Hyp$ are also `stochastic' 
elements\footnote{In a PAC-Bayesian context, we will merely consider weighted sums of the elements in $\Hyp$,
rather than assuming a truly Bayesian setup.}, possessing some notion of likelihood, 
say $Q:\Hyp\rightarrow[0,1]$ such that $\int_\Hyp Q(h_r) dh = 1$.
Consider at first the case where $\Hyp$ is finite, 
we are interested in what happens on functions $E_Q[h_r(x)]$, which is defined as 
\begin{equation}
	E_Q\left[h_r(x)\right] = \sum_{h_r\in\Hyp}  h_r(x) Q(h_r). 
	\label{eq.EQ}
\end{equation}
If $|\Hyp|$ is infinite, then the sum can be replaced by an integral as usual, or 
$E_Q\left[h_r(x)\right] = \int_{\Hyp}  h_r(x) Q(h_r) dh_r$. 
In the analysis we will assume $|\Hyp|<\infty$ in order to avoid technical issues.
Note that this is not quite a regular (well-known) expectation $\E[\cdot]$ as before.
Now let the Kullback-Leibler distance be defined for each $0<p,q<1$ be defined as 
	$\KL(q,p) = q\log\frac{q}{p} + (1-q)\log\frac{1-q}{1-p}$,
where $\log(\cdot)$ denote the natural logarithm. Let the function $P:\Hyp\rightarrow[0,1]$ be the prior weighting function over $\Hyp$. If $Q:\Hyp\rightarrow[0,1]$ and $P:\Hyp\rightarrow[0,1]$ are two functions, we extend the definition as 
\begin{equation} 
	\KL(Q,P) = \sum_{h_r\in_\Hyp} Q(h_r) \log\frac{Q(h_r)}{P(h_r)}.	
	\label{eq.KL2}
\end{equation}
We state the PAC-Bayes theorem as in \cite{maurer2004note}:
\begin{theorem}
	For $\delta>0$ and for $n\geq 8$, we have that with probability exceeding $1-\delta$ we have that 
	for all $Q:\Hyp\rightarrow[0,1]$ the following inequality holds:
	\begin{equation}
		\KL\Big( E_Q[\risk_n(h_r)], E_Q[\risk(h_r)] \Big) \leq \frac{\KL(Q,P) + \log\frac{1}{\delta} + \log(2\sqrt{n})}{n}.
		\label{eq.pacbayes}
	\end{equation}
\end{theorem}
Specifically, this holds for a $Q_n$ found by an algorithm based on the $n$ i.i.d. observations.
Note that this result is currently the most tight inequality, refining the ideas presented in \cite{mcallester1999pac}.
While till date most applications are found in the context of supervised learning, 
we will argue in the following that this theorem finds a `natural' application towards unsupervised learning.

\subsection{An Application of PAC-Bayes Towards Clustering}

In what follows, assume that 
the $n$ i.i.d. samples $\{X_i\}_{i=1}^n$ take values in a bounded set in $S\subset\R^d$ for a given $d\in\N$.
In order to use the PAC-Bayes result to the generic application of clustering, we need 
to specify the loss function $\ell:\R^d\rightarrow[0,1]$ of interest. 
A `cluster', represented as an indicator function $h:\R^d\rightarrow\{0,1\}$, is understood here as a member 
of a user-specified set of indicator functions $\Hyp= \left\{h:\R^d\rightarrow\{0,1\}\right\}$. Formally,
one defines for a set $c\subset\R^d$
\begin{equation}
	h_c(x) = I(x\in c) =   
	\begin{cases}
		1 &x\in c  \\
		0 & x\not\in c.
	\end{cases}
	\label{eq.cluster}
\end{equation}
Now, we look a bit closer at what the term $E_Q[\risk(h_c)]$ represents in this context. 
\begin{equation}
	{ E_Q[\risk(h_c)] = \sum_{h_c\in\Hyp}\prob(X\in c) Q(h_c) \\
	= \E\left[\sum_{h_c\in\Hyp} h_c(X)Q(h_c) \right]},
	\label{eq.explanatory}
\end{equation}
where the second equality holds by linearity of the expectation, and 
where $\prob$ denotes the probability rules underlying the data.
Consequently, the term $E_Q[\risk(h)]$ characterizes how well $Q$ 
{\em aligns} with the distribution underlying the data. 
Assume that the $\Hyp$ is designed such that all 
sets $c$ corresponding to a $h_c\in\Hyp$ (i) cover the space $S$ and (ii) are disjunct.

The function $P:\Hyp\rightarrow[0,1]$ is the prior weighting function  (think of it as a `prior distribution' over $\Hyp$).
In general, it is up to the user in a specific application to decide how to design $(\Hyp,P)$:
it is good practice to make it equally likely for each hypothesis $h\in\Hyp$ to explain the data by itself,
- suggesting a uniform prior $P$ over this set $\Hyp$-  
while the result should be useful for the application in mind.
Assume for example that all probability mass (underlying the 
samples) concentrates in the set corresponding with a single $h_c$, 
and $Q(h_c)=I(i=j)$, then this measure equals 1.
On the other hand, if all samples are equally  distributed over the $|\Hyp|$ sets $h_c\in\Hyp$, 
the measure equals $\frac{1}{|\Hyp|}$.
This motivates the naming of $E_Q[\risk(h)]$ as the {\em explanatory power} of $(\Hyp,Q)$. 
Specifically, if $\Hyp= \{I(x\in[-1,1]^d)\}$, the explanatory power of $(\Hyp,Q)$ is 1, but 
it however is not very useful, surprising nor {\em falsifiable}. 


We argue that this PAC-Bayesian interpretation to clustering is often `natural' because of three reasons.
	(i) The present analysis does not need to recover the density function underlying the data, 
	a feature which is highly desirable if working with high-dimensional data.  
	(ii) The set of `underlying' clusters is not recovered exactly, nor assumed to exists in reality.
	The actual stochastic rules underlying the observed data only say how well the hypothesis clustering `explains' the data.	
	When dealing with data arising from complex processes the assumption of a `true clustering' is often an oversimplification.
	(iii) The characterization of performance of the found rule $Q_n$ in terms of its deviation from the prior $P$ is desirable  if 
	clustering is meant for looking for `consistent' irregularities. Specifically, if the result $Q_n$ is 
	not what we (more or less) expected before seeing the data, substantial empirical evidence should be presented 
	motivating this property. 
Those reasons differentiate the approach substantially from approaches based on density estimation, 
or on mixtures of distributions. 
Remark that this description of explanatory power is strongly related to the ideas presented in \cite{taylor2007}.
The following clustering algorithm is then motivated by application of the PAC-Bayesian theory:
\begin{equation}
	Q_n =\argmin_Q E_Q[\risk_n(h)] \mbox{\ s.t. \ } \KL(Q,P)\leq \omega_n,
	\label{eq.clustering}
\end{equation}
where $\omega_n>0$. 
This objective is also motivated  from an information theoretical approach to clustering, as e.g. in \cite{banerjee2005clustering}.

\subsection{An Application of PAC-Bayes Towards Pairwise Clustering}

Now we explain how the above insights lead to an analysis of the pairwise clustering setup. 
Let again $\X$ and $\Y$ denote respectively the two domains of interest in which pairwise observations $(x,y)$ are made.
A first approach would be  to rephrase the pairwise clustering 
problem as a standard clustering approach, where instead of the class of indicator functions 
$\Hyp_f\subset\{f:\X\rightarrow[0,1]\}$ in the first domain,  
one studies the cross-product of this class 
with the class of indicator functions in the other domain $\Hyp^{f,g} = \Hyp_f\times \Hyp_g$, or  
\begin{equation}
	\Hyp^{f,g}	\subset \Big\{h=(f_h,g_h)\ \Big| \ f_h:\X\rightarrow[0,1], g_h:\Y\rightarrow[0,1]\Big\}.
	\label{eq.hyp2}
\end{equation}
However, the reasoning in the introduction suggests another route.
To see this, we formalize the intuition of the pairwise observation $(x,y)$ being a target for prediction:
(i) let $z\in\X$ represent the part  of a sample $x=(z,y)$ which might be used to predict (a property) of the (unobserved) $y\in\Y$; and/or  
(ii) given $y\in\Y$, predict (a property) of the corresponding (unobserved) $z\in\X$. 
Given a set $\Hyp^{f,g}$: the knowledge of the `cluster' to which $X$ belongs, 
will be used to predict the cluster memberships of the corresponding $y$.

We will say that $f_h$ {\em explains} $z\in\X$ if $f_h(z)=1$, and similarly that $g_h$ {\em explains} $y\in\Y$ if $g_h(y)=1$.
In an ideal case, one would be able to associate exactly one distinct $f_h\in\Hyp_f$ to every $g_h\in\Hyp_g$ (i.e. describe a permutation). 
As such, one could predict the cluster $g_h$ containing $y$ corresponding to a given $z$.
In the worst case, the choice of $g$ that explains $y$ is independent of $z$ being explained by $f$.
The pairwise clustering setup however differs from such a multi-class classification (structured output prediction)
task as it is essentially symmetric: a given $z$ is used to predict (cluster membership of) the corresponding $y$, and 
a given $y$ is used to predict (cluster memberships of) the corresponding $x$.
Now, a pairwise cluster $h=(f,g)\in\Hyp^{f,g}$ was useful for a sample $(z,y)\in\X\times\Y$ if 
$f(z) = g(y)$. Alternatively, a pairwise cluster $c=(f,g)$ {\em contradicts} a sample 
if $f(z)\neq g(y)$. This motivates the following risk function
\begin{equation}
	\begin{cases}
		\risk_n(h) = \frac{1}{n} \sum_{i=1}^n I(f_h(Z_i)\neq g_h(Y_i))\\
		\risk(h) = \prob\left(f_h(Z) \neq g_h(Y)\right),		
	\end{cases}
	\label{eq.risk.pwca}
\end{equation}
defined again in an `empirical' and an `actual' flavor. 
This definition measures how many (for how large a probability mass) 
datapoints are contradicted by a pairwise cluster $h = (f_h,g_h)$.
Now the term $E_Q[\risk(h)]$ becomes
\begin{equation}
	{ E_Q[\risk(h)] = \sum_{h\in\Hyp^{f,g}}\prob\left(f_h(Z)\neq g_h(Y)\right) Q(h)},
	\label{eq.EQrisk}
\end{equation}
which basically captures how many mistakes are made when focussing on the subset of $\Hyp^{f,g}$ 
as directed by $Q$.
This motivates the following practical approach:
(i) given a dataset $\{X_i=(Z_i,Y_i)\}_{i=1}^n$, with the elements taking values in $\X\times\Y$, 
and (ii) a a set $\Hyp^{f,g}$ of pairwise clusters represented as  $h=(f,g)$, 
and a `prior' weighting function $P:\Hyp^{f,g}\rightarrow[0,1]$, then
we aim to find a new weighting function $Q_n:\Hyp^{f,g}\rightarrow[0,1]$ which is not too different from $P$,
and which aligns well with the probability rules underlying the data as 
\begin{equation}
	Q_\ast = \argmin_Q E_Q(\risk(h_c)) \mbox{\ s.t. \ } \KL(Q,P)\leq \omega,
	\label{eq.pwca}
\end{equation}
where $\omega>0$.
The PAC-Bayes theorem now guarantees that this problem is approximatively solved based on the data as
\begin{equation}
	Q_n' = \argmin_Q E_Q(\risk_n(h_c)) \mbox{\ s.t. \ } {\KL}(Q,P)\leq \omega,
	\label{eq.pwca2}
\end{equation}
where $\omega>0$.
The resulting $Q'_n$ will emphasize the pairwise clusters which are most often consistent with the data.
Here we have a natural trade-off between specificity and accuracy, regulated by $\omega_n$.
If $\omega_n$ were small, the solution $Q'_n$ cannot deviate from the uniform distributions over all pairwise clusters in $\Hyp^{f,g}$,
but then many different pairwise clusters will contradict on different samples, leading in turn to low explanatory power.
On the other hand, allowing for arbitrary $Q'_n$ will explain the individual samples fairly well (allowing a single pairwise cluster per sample), 
but the PAC-Bayesian result will not guarantee accuracy of the result anymore.

\begin{figure}[ht]
         \center
	\begin{tabular}{cc}
		\subfigure{\includegraphics[width=0.18\textwidth]{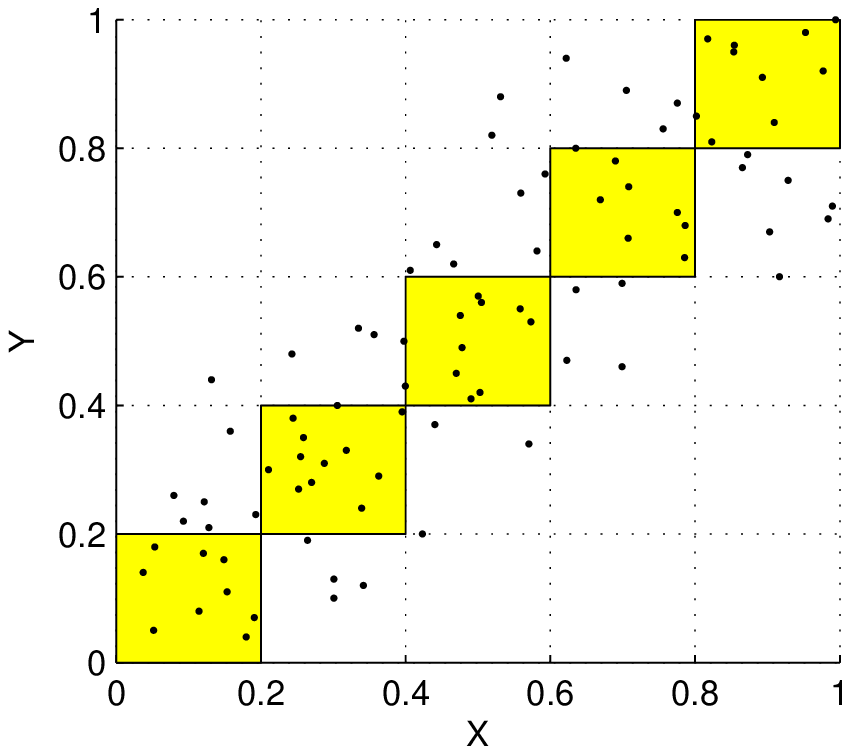}} &
		\subfigure{\includegraphics[width=0.18\textwidth]{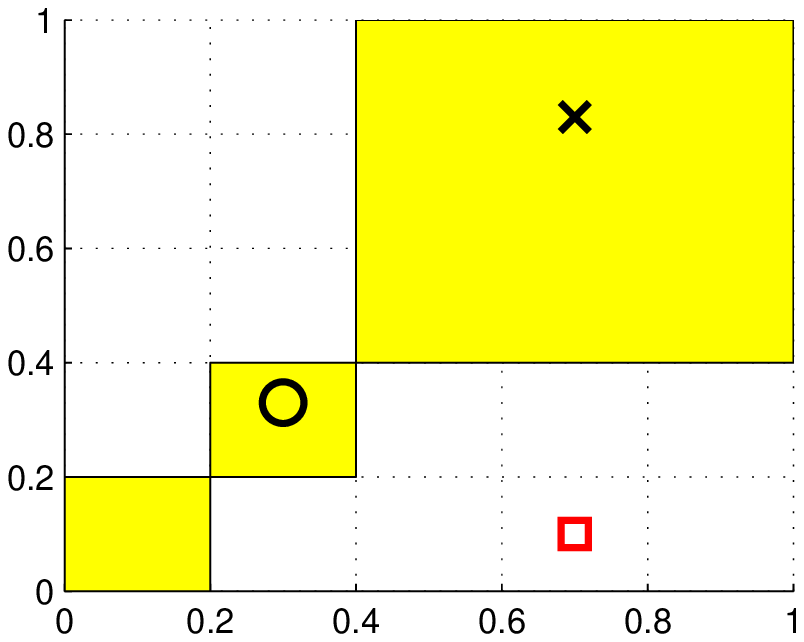}}
	\end{tabular}
	\caption{
		Schematic representation of all pairwise clusters in a hypothesis space $\Hyp$ 
		based on the 5 disjunct intervals $d+[0, 0.2]$ in either domain (dotted lines).
		The dots $(X,Y)\in\R\times\R$ represent samples from an underlying distribution. 
		Suppose the different hypothesis 
		can be  factorized as $h_c=(f,g)$, where $f:\R\rightarrow[0,1]$ and $g:\R\rightarrow[0,1]$,
		being the corresponding indicator functions in either domain.
		This means that there are 25 possible different pairwise clusters $h_c$ (dotted squares), or $|\Hyp^{f,g}|=25$, 
		(a) about $70\%$ of the observations (dots) do not contradict the 5 pairwise clusters (yellow squares) simultaneously;
		(b) Only one sample (`$\square$') contradicts the shown pairwise cluster $h_c$ (yellow squares),
		while the other two (`$\circ$' and `$\times$') are consistent with $h_c$.
		}
	\label{grid2}
\end{figure}

We now express the `regularization term' $\KL(Q,P)$ in a more convenient form.
\begin{proposition}[Bound to K.-L. Divergence]
	Assume $|\Hyp|<\infty$ and $P(h) = \frac{1}{|\Hyp|}$ for all $h\in\Hyp$, then
	\begin{equation}
		\KL(Q,P) 
		\leq \log \sum_{h\in\Hyp} Q^2(h) + \log(|\Hyp|). 
		\label{eq.kl}
	\end{equation}
\end{proposition}
This is a consequence of the following inequality on the entropy of a vector $p\in]0,1[^d$ with $1^Tp=1$ 
\begin{equation}
	\entropy(p) = \sum_{i=1} p_d\log(p_d)  \leq \log\left(\sum_{i=1}^d p_i^2\right),
	\label{eq.entropy}
\end{equation}
by application of Jensen's inequality.
Let $s^Q\in[0,1]^{|\Hyp|}$ be a vector representing the function $Q$ where $s_i^Q = Q(h_i)$ 
(enumerating the different elements  $h_i\in\Hyp$), then
\begin{equation}{
	s_n^Q = \argmin_{s^Q\geq 0_n\sum_i s^Q_i=1} 
	\|s^Q\|_2 \mbox{ \ s.t. \ }E_Q[\risk_n(h)] = 0.}
	\label{eq.pwca3}
\end{equation}
implementing the socalled {\em realizable} case (as in the theory of Support Vector Machines).
The optimal solution $Q_n$ will try to find as many pairwise clusters as possible which are not contradicting the given data. We illustrate this notion in figure \ref{grid2}. In the ideal case, all observations are explained. In more realistic cases, merely a few pairwise clusters are found (i.e., the set $\{h\in\Hyp: Q(h)>0\}$ contains only a few elements). 

We extend this model to account for infinite $\Hyp$, defined as $h=(\delta_z,\delta_y)$
for each $(z,y)\in\X\times\Y$, and where $\delta_x$ denotes the Dirac delta.
When extending the formulation in order to deal with infinite hypothesis spaces $\Hyp^{f,g}$,
we replace vectors $s_Q$ by  functions $Q:\Hyp\rightarrow\R^+$, 
which (for convenience) are assumed to be elements of a Hilbert spaces $\mathbf{H}$.
This space is equipped with a corresponding inner-product (reproducing kernel) 
$k:\mathbf{H}\times\mathbf{H}\rightarrow\R$, implicitly defining $\Hyp$ and $P$. 
Note that $Q(h)\geq 0$ for all $h\in\Hyp$, and $\int_\Hyp Q(h)dh=1$.
This motivates the replacement of the term $\KL(Q,P)$ by $\|Q\|_{\mathbb{H}}$.
As such (\ref{eq.pwca2}) is equivalent (up to normalization) to
\begin{equation}{
	Q''_n = \argmin_{Q} 
	\|Q\|_\mathbf{H} \mbox{ \ s.t. \ }E_Q[\risk_n(h)] = 0.}
	\label{eq.pwca4}
\end{equation}
where $Q''_n(h)\geq 0$ for all $h\in\Hyp$, and $ \int_{\Hyp} Q''_n(h) dh=1$.
Note that for the majority of pairwise clusters no data is sampled contradicting the 
cluster, and a smooth transition of $Q$ inbetween the sample becomes possible.
In the remainder we will assume the relevant Hilbert space $\mathbf{H}$
can be decomposed additively uniquely as $\mathbf{H}_\X \otimes \mathbf{H}_\Y$, and the norm of a function $Q$
can then be written as $\|Q\|^2_\mathbf{H} = \|F\|^2_{\mathbf{H}_\X} + \|G\|^2_{\mathbf{H}_\Y}$.
Assume $\Hyp^{f,g}$ contains all pairwise clusters $h=(\delta_z,\delta_y)$ for all $(z,y)\in\X\times\Y$ and $\delta$ the Dirac delta.
Under the assumtion no ties occur in the data, problem (\ref{eq.pwca4}) is 
\begin{equation}
	(F_n,G_n) = \argmin_{F,G}  \|F\|^2_{\mathbf{H}_\X} + \|G\|^2_{\mathbf{H}_\Y} 
	\mbox{ \ s.t. \ } F_i = G_i, \ \forall i=1,\dots,n.
	\label{eq.pwca4}
\end{equation}
enforcing that $F(h) = G(h)$ for all $h\in\Hyp^{f,g}$,
and enforcing again that $F(h)\geq 0$ for all $h\in\Hyp^{f,g}$ as well as that $\int_{\Hyp^{f,g}} F(h) dh=1$.
Here $F_i=F(\delta_{Z_i})$ and $G_i(\delta_{Y_i}) = Q((\delta_{Z_i},\delta_{Y_i}))$ for all $i=1,\dots,n$.
The next section shows how to solve this problem, relaxing the (in)equality constraints.

\section{Kernel PairWise Component Analysis}

\subsection{PWCA for paired Observations}

This section studies how the learning problem (\ref{eq.pwca4}) is solved (approximatively) by an efficient algorithm.
Let $X^a = (X_1^T, \dots, X_n^T)^T \in \mathbb{R}^{\ell\times m}$ 
and $Y^b=(Y_1^T, \dots, Y_n^T)^T \in \mathbb{R}^{\ell\times n}$ be matrices where $\ell$ is the number of samples 
and $m,n$ are the number of attributes/features for the first and second representation respectively. 
The functions $Q$ are parametrised as  $F_{\mathbf{v}_c}(z)=\mathbf{v}_c^Tz$ and $G_{\mathbf{w}_c}(y)=\mathbf{w}_c^Ty$.
The inequalities $Q(h)\geq 0$ are enforced by representing this as 
$Q(h)=F^2(f)=G^2(h)$ for all $h\in\Hyp^{f,g}$.
This is imposed by enforcing
$c_i = \sqrt{Q((\delta_{Z_i},\delta_{Y_i}))} =  F(\delta_{Y_i})= G(\delta_{Y_i})$, 
then $\int_\Hyp Q(h)dh=1$ is enforced by imposing the constraint $\mathbf{c}'\mathbf{c}=1$ (similarly, maximizing $\mathbf{c}'\mathbf{c}$).
As such (\ref{eq.pwca2}) becomes
\begin{equation}
\max_{\mathbf{c}\in\R^\ell, \mathbf{v}_c\in\mathbb{R}^m, \mathbf{w}_c\in\mathbb{R}^n} 
\mathbf{c}'\mathbf{c} - \gamma (\mathbf{w}_c' \mathbf{w}_c + \mathbf{v}_c'\mathbf{v}_c),
\label{eqn:main}
\end{equation}
where $A'$ is the transpose of matrix, or vector, $A$ and such that 
$\mathbf{c}_i = X_{a,{i}}\mathbf{w}_c = Y_{b,{i}} \mathbf{v}_c,$ for $i=1,\dots,\ell$. 
Associating Lagrange multipliers $\alpha_i,\beta_i$ to each of the $\ell$ constraints gives the following Lagrangian
\begin{eqnarray}
\label{eqn:lag}
\mathcal{L} & = &\frac{1}{2} \mathbf {c}'\mathbf{c} - \frac{\gamma}{2} (\mathbf{w}_c'\mathbf{w}_c + \mathbf{v}_c'\mathbf{v}_c) 
 - \bm{\alpha}'(\mathbf{c} - X_a \mathbf{w}_c) - \bm{\beta}'(\mathbf{c} - Y_b \mathbf{v}_c).
\end{eqnarray}
Taking derivatives of equation (\ref{eqn:lag}) with respect to $\mathbf{w}_c, \mathbf{v}_c, \mathbf{c}$ and setting to zero give the following conditions for optimality as
\begin{equation*}
	\dfrac{\partial \mathcal{L}}{\partial \mathbf{w}_c} =\mathbf{0} \rightarrow
	\mathbf{w}_c =\frac{1}{\gamma} X_a' \bm{\alpha}, \
	\dfrac{\partial \mathcal{L}}{\partial \mathbf{v}_c} =\mathbf{0} \rightarrow
	\mathbf{v}_c = \frac{1}{\gamma}Y_b' \bm{\beta}, \ 
	\dfrac{\partial \mathcal{L}}{\partial \mathbf{c}} =\mathbf{0} \rightarrow
	\mathbf{c} = (\bm{\alpha}+\bm{\beta}).
\end{equation*}
Setting back into the optimisation in equation (\ref{eqn:main}) gives the following dual problem
\[
\max_{\bm{\alpha}\in\mathbb{R}^\ell,\bm{\beta}\in\mathbb{R}^\ell} \mathcal{J} = \frac{1}{2}(\bm{\alpha}+\bm{\beta})'(\bm{\alpha}+\bm{\beta}) - \frac{1}{2 \gamma} (\bm{\alpha}'K_a\bm{\alpha} + \bm{\beta}'K_b\bm{\beta}),
\]
where $K_a = X_aX_a'$ and $K_b = Y_bY_b'$ are the kernel matrices. Taking derivatives and setting to zero shows that $\mathcal{J}$ achieves a (local) optimum when 
\begin{eqnarray}
\label{eq:sol}
	\dfrac{\partial \mathcal{J}}{\partial \bm{\alpha}} =\mathbf{0} &\rightarrow & \gamma(\bm{\alpha}+\bm{\beta}) =  K_a \bm{\alpha} \\
	\dfrac{\partial \mathcal{J}}{\partial \bm{\beta}} =\mathbf{0}&\rightarrow & \gamma(\bm{\alpha}+\bm{\beta}) = K_b \bm{\beta}. 
\nonumber
\end{eqnarray}
We are able to observe that at optimum $K_a \bm{\alpha} = K_b \bm{\beta}$, which illustrates a direct relationship to KCCA condition. Due to limited space we do not explore the relationship to KCCA within the scope of this manuscript.  Equation (\ref{eq:sol}) can be rewritten as
\begin{equation}
\label{eqn:mainblock}
\begin{bmatrix} 
K_a & 0_\ell \\ 
0_\ell & K_b
\end{bmatrix}
\begin{bmatrix} 
\bm{\alpha} \\ 
\bm{\beta}
\end{bmatrix}
= {\gamma}
\begin{bmatrix} 
I_\ell & I_\ell \\ 
I_\ell & I_\ell
\end{bmatrix}
\begin{bmatrix} 
\bm{\alpha} \\ 
\bm{\beta}
\end{bmatrix},
\end{equation}
where $I_\ell$ is the identity matrix and $0_\ell$ is a matrix of zeros, both of size $\ell\times\ell$. This equation may be solved as a generalized eigenvalue problem in the form of $A\mathbf{x} = \lambda B \mathbf{x}$. Alternatively, we observe that by setting $\bm{\beta} = \left(\frac{1}{\gamma} K_a - I\right)\bm{\alpha}$, we can express
$\frac{1}{\gamma} K_a \bm{\alpha} =  \frac{1}{\gamma^2} K_b K_a \bm{\alpha} - \frac{1}{\gamma} K_b\bm{\alpha}$,
which results in the following generalized eigenvalue problem for $\bm{\alpha}$
\begin{equation}
\label{eqn:meig}
 K_bK_a\bm{\alpha} = \gamma \left(K_a + K_b\right) \bm{\alpha},
\end{equation}
and by setting $R$ to be the Cholesky decomposition of $K_bK_a$ such that $K_bK_a = RR'$ we obtain the following symmetric eigenvalue problem
\begin{equation*}
I_\ell \bm{\alpha} = \gamma R^{-1}\left(K_a + K_b\right)R^{{-1}'} \bm{\alpha}.
\end{equation*}
It may be necessary to regularize equation (\ref{eqn:mainblock}) with some small value $\tau$ on the diagonal. This will result in our optimisation being rewritten as
\begin{equation*}
\begin{bmatrix} 
K_a & 0_\ell \\ 
0_\ell & K_b
\end{bmatrix}
\begin{bmatrix} 
\bm{\alpha} \\ 
\bm{\beta}
\end{bmatrix}
= {\gamma}
\begin{bmatrix} 
I_\ell(1 + \tau) & I_\ell \\ 
I_\ell & I_\ell (1+ \tau)
\end{bmatrix}
\begin{bmatrix} 
\bm{\alpha} \\ 
\bm{\beta}
\end{bmatrix}.
\end{equation*}
Furthermore, the above eigenvalue problem can be written as
$\bm{\beta} = \left(\frac{1}{\gamma} K_a - \tau I_\ell \right)\bm{\alpha}$
and
\begin{eqnarray*}
K_bK_a\bm{\alpha} = \gamma^2 (I_\ell- \tau^2 I_\ell)\bm{\alpha} + \gamma (\tau I_\ell K_a + \tau I_\ell K_b )\bm{\alpha},
\end{eqnarray*}
which can be solved as a quadratic eigenvalue problem.
It follows from the conditions for optimality that a new sample $(\mathbf{\bar{x}}_a, \mathbf{\bar{y}}_b)$ can be projected in the learnt semantic space by the functions
\[\begin{cases}
	F(\mathbf{\bar{x}}_a) =  \mathbf{w}_c' \mathbf{\bar{x}}_a =\frac{1}{\gamma} \bm{\alpha}'K_a(\mathbf{x}_{a},\mathbf{\bar{x}}_a) , \\ 
	G(\mathbf{\bar{y}}_b) =\mathbf{v}_c' \mathbf{\bar{y}}_b  = \frac{1}{\gamma}\bm{\beta} 'K_b(\mathbf{y}_{b},\mathbf{\bar{y}}_b). 
\end{cases}\] 
Then it is also reasonable to assign the sample $(\mathbf{\bar{x}}_a,\mathbf{\bar{y}}_b)$ to the cluster $(1,\ldots, \ell)$ which has highest (absolute) factors $\left| F(\mathbf{\bar{x}}_a)\right|_{1}^{\ell}$ and $\left|G(\mathbf{\bar{y}}_b) \right|_{1}^{\ell}$ respectively.

\subsection{PWCA for  Multiview Observations}

In this section we generalize our methodology to multiple views. Expressing optimization in equation (\ref{eqn:main}) for three sources gives
\begin{equation}
\label{eqn:mway}
\max_{\mathbf{c}\in\mathbb{R}^\ell, \mathbf{w}_c\in\mathbb{R}^m, \mathbf{v}_c\in\mathbb{R}^n, \mathbf{z}_c\in\mathbb{R}^s} 
\frac{1}{2} \mathbf{c}'\mathbf{c} - \frac{\gamma}{2} (\mathbf{w}_c' \mathbf{w}_c + \mathbf{v}_c'\mathbf{v}_c + \mathbf{z}_c'\mathbf{z}_c),
\end{equation}
such that $c_i = X_{a,i} \mathbf{w}_c = X_{b,i} \mathbf{v}_c = X_{c,i}\mathbf{z}_c ,$ for $i=1,\dots,\ell$. Taking derivatives of equation (\ref{eqn:mway}) with respect to $\mathbf{w}_c, \mathbf{v}_c , \mathbf{z}_c, \mathbf{c}$ and setting to zero will give the conditions for optimality. Substituting these conditions back into equation (\ref{eqn:mway}) gives the following dual problem
\begin{eqnarray*}
\max_{\bm{\alpha}\in\mathbb{R}^\ell,\bm{\beta}\in\mathbb{R}^\ell,\bm{\nu}\in\mathbb{R}^\ell} \mathcal{J} &= &\frac{1}{2}(\bm{\alpha}+\bm{\beta}+\bm{\nu})'(\bm{\alpha}+\bm{\beta}+\bm{\nu}) - \frac{1}{2 \gamma} (\bm{\alpha}'K_a\bm{\alpha} + \bm{\beta}'K_b\bm{\beta} + \bm{\nu}'K_c\bm{\nu}),
\end{eqnarray*}
where $K_a = X_aX_a'$, $K_b = X_bX_b'$ and $K_c = X_cX_c'$ are the kernel matrices. Taking derivatives and setting to zero shows that $\mathcal{J}$ achieves a (local) optimum when 
\begin{equation*}
	\dfrac{\partial \mathcal{J}}{\partial \bm{\alpha}} = \mathbf{0} \rightarrow  \gamma(\bm{\alpha}+\bm{\beta}+\bm{\nu}) =  K_a \bm{\alpha}, \
	\dfrac{\partial \mathcal{J}}{\partial \bm{\beta}} =\mathbf{0} \rightarrow  \gamma(\bm{\alpha}+\bm{\beta}+\bm{\nu}) = K_b \bm{\beta}, \
	\dfrac{\partial \mathcal{J}}{\partial \bm{\nu}} =\mathbf{0} \rightarrow \gamma(\bm{\alpha}+\bm{\beta}+\bm{\nu}) = K_c \bm{\nu}.
\end{equation*}
which can be rewritten as
\begin{equation*}
\begin{bmatrix} 
K_a & 0_\ell & 0_\ell \\ 
0_\ell & K_b & 0_\ell \\ 
0_\ell & 0_\ell & K_c
\end{bmatrix}
\begin{bmatrix} 
\bm{\alpha} \\ 
\bm{\beta} \\
\bm{\nu}
\end{bmatrix}
= {\gamma}
\begin{bmatrix} 
I_\ell & I_\ell  & I_\ell\\ 
I_\ell & I_\ell  & I_\ell\\
I_\ell & I_\ell & I_\ell
\end{bmatrix}
\begin{bmatrix} 
\bm{\alpha} \\ 
\bm{\beta} \\
\bm{\nu}
\end{bmatrix},
\end{equation*}
where again $I_\ell$ is the identity matrix and $0_\ell$ is a matrix of zeros, both of size $\ell\times\ell$. 
Therefore, without loss of generality, we can extend this to multiple $\mathbf{i}=1,\ldots,s$ views, where $s \geq 2$, similarly to the previously proposed multi-view extension for CCA by \cite{BachJordan}, such that
\begin{equation*}
\begin{bmatrix} 
K_1 & \ldots & 0_\ell \\ 
\vdots & \ddots & \vdots  \\ 
0_\ell & \ldots & K_{s}
\end{bmatrix}
\begin{bmatrix} 
\bm{\alpha}_1 \\ 
\vdots \\
\bm{\alpha}_s
\end{bmatrix}
= {\gamma}
\begin{bmatrix} 
I_\ell & \ldots  & I_\ell\\ 
\vdots & \ddots  & \vdots\\
I_\ell & \ldots & I_\ell
\end{bmatrix}
\begin{bmatrix} 
\bm{\alpha}_1 \\ 
\vdots \\
\bm{\alpha}_s
\end{bmatrix}.
\end{equation*}
This equation may be solved as a generalized eigenvalue problem in the form of $A\mathbf{x} = \lambda B \mathbf{x}$.

\section{Experiments of PWCA on Europal}

We proceed to compare PWCA to KCCA for a mate-retrieval task \cite{Vino02,Li06, Fortuna08,Guezouli2010}, i.e. given a document query $\mathbf{q}_i$ in language $x$ to retrieve the (exact) matching document in the paired language $y$. For this purpose we use the multi-lingual Europal dataset \cite{Koehn}, which has a total of 11968 aligned documents. We use the following eight languages with the number of features/words in brackets; da - Danish (78720), de - German (153499), en - English (60369), es - Spanish (171821), it - Italian (66548), nl - Dutch (105318), pt - Portuguese (66922) and sv - Swedish (51116). We use linear kernels throughout and arbitrarily set the regularization parameter to $\tau = 0.01$ for both methods. Finally, the performance is evaluated using Average Precision (AP) \cite{Turpin} which is computed as $AP = \frac{1}{\ell} \sum_{i=1}^\ell \frac{1}{I_i}$ where $I_i$ is the rank location of the exact paired document for query document $\mathbf{q}_i$. Therefore $AP = 0.5$ indicates that the paired document is on average situated at location $I = 2$. We select the rank by sorting the, absolute, inner products values of $F(\mathbf{q}_i)'G(\mathbf{y}_j)$ (as well as for $F(\mathbf{x}_i)'G(\mathbf{q}_j)$) for all possible paired test documents, i.e. we rank the retrieved documents according to their similarity (in the learnt space) with our query. In our experiments we use the CCA formulation as proposed by \cite{BachJordan} for both pair- and multi-view.

In the first of our two experiments, for each pairing combination of languages, we randomly select 500 paired-documents for training and 5000 for testing. The analysis has been repeated 10 times and averaged across. The results given in table \ref{tab:2way} are the AP averaged across of all possible language-pair combinations for the language indicated in the column (i.e. column {\it da} is the average of all the language pairing with {\it da - xx}). We are able to observe that PWCA is able to perform, on average, on a par with KCCA. The mean AP across all languages for KCCA is 0.4435 whereas for PWCA it is 0.4459.

\begin{table}[ht]
\caption{ We compare KCCA and PWCA on a bilingual mate-retrieval task (see text for language abbreviation). The reported results are the AP for retrieving the exact paired document in another language, averaged across all possible language-pair combination for the language indicated in the column. The results are averaged over 10 repeats of the analysis.}
\begin{center}
\begin{tabular}{|l|c|c|c|c|c|c|c|c|}
\hline
 & da & de & en & es  & it & nl & pt & sv\\
\hline
KCCA & 0.4174 & 0.3839 & {\bf 0.4979} & 0.4243&  {\bf 0.4572} & 0.4023 & {\bf 0.4939} & 0.4714\\
PWCA & {\bf 0.4294} & {\bf 0.4416} & 0.4747 & {\bf 0.4344}&  0.4368 & {\bf 0.4111} & 0.4679 & {\bf 0.4716}\\
\hline
\end{tabular}
\end{center}
\label{tab:2way}
\end{table}%

In the second experiment we extend the previous analysis to a trilingual mate-retrieval task, i.e. we train on an aligned document corpus from three languages whereas during testing we compute the mean average precision of all the individual pair-wise mate-retrieval tasks (of the three languages). In other words, we train on the trilingual alignment of {\it da-de-en} while we test the query retrieval on the bilingual task of {\it da-de, da-en, de-en}. In this experiment we randomly select 500 tripartite-documents for training and 2000 for testing. Due to increased complexity we only repeat the analysis, for each 3 language combination, once. The results given in table \ref{tab:3way}, as in the previous table, are the mean average precision for the language stated in the column and all its possible tripartite combinations (without repetition, i.e. for example; {\it da-da-en} is not be allowed). We are clearly able to see the improvement gained by PWCA over KCCA despite increasing the training alignment complexity. Furthermore, not only did the added aligned language not hinder the mate retrieval task, it improved performance as visible when comparing table \ref{tab:2way} with table \ref{tab:3way}.
\begin{table}[ht]
\caption{ We compare KCCA and PWCA on a trilingual mate-retrieval task (see text for language abbreviation). The reported results are the mean average precision for retrieving the exact paired document in another language for all possible tripartite combinations of the language stated in the column (without repetition) for training.}
\begin{center}
\begin{tabular}{|l|c|c|c|c|c|c|c|c|}
\hline
 & da & de & en & es& it & nl & pt & sv\\
\hline
KCCA & 0.3687 &  0.3290 & 0.3930 & 0.3742& 0.3792 & 0.3501 & 0.3917 & 0.3909\\
PWCA & \bf 0.5407 & \bf 0.5155 & \bf 0.5427 & \bf 0.5394& \bf 0.5310 & \bf 0.5246 & \bf 0.5406 & \bf 0.5504 \\
\hline
\end{tabular}
\end{center}
\label{tab:3way}
\end{table}%

CCA (and KCCA) does not seek to maintain any pre-existing structure within the views while seeking to maximise correlation across the views. This aspect that may lead to over-fitting when having multiple views, PWCA addresses this by directly seeking to maintain internal structure by trying to find as many pairwise (or n-wise) clusters as possible which do not contradict the given data. We hypothesis that the PWCA performance improvement is a direct result of the clustering condition.

\section{Discussion}
This study presented a novel learning paradigm and corresponding algorithm that aims at finding structure (pairwise clusters) in paired (multi-view) observations.
A case study on bilingual and trilingual mate-retrieval task, and a motivation using the PAC-Bayesian results are given. 
While this paper described a theoretical as well as applied proof of concept, many issues including efficiency, out-of-sample extensions and relations to other techniques remain.

%

%





%



\end{document}